%% file: PerAwareCity.tex
\documentclass[conference]{IEEEtran}
\IEEEoverridecommandlockouts
\usepackage{cite}
\usepackage{amsmath,amssymb,amsfonts}
\usepackage{algorithmic}
\usepackage{graphicx}
\usepackage{textcomp}
\usepackage{xcolor}
\usepackage{comment}

\usepackage{latexsym}
\usepackage{multirow}
\usepackage{url}
\usepackage{here}
\usepackage{makecell}

\usepackage{tabularx}
\usepackage{siunitx}

\def\BibTeX{{\rm B\kern-.05em{\sc i\kern-.025em b}\kern-.08em
    T\kern-.1667em\lower.7ex\hbox{E}\kern-.125emX}}
\begin{document}

\title{Bus Ridership Prediction with Time Section, Weather, and Ridership Trend Aware \\ Multiple LSTM}

\author{
    \IEEEauthorblockN{Tatsuya Yamamura}
    \IEEEauthorblockA{
        \textit{Graduate School of Science and Technology} \\ 
        \textit{Nara Institute of Science and Technology} \\ 
        Nara, Japan \\
        tatsuya.yamamura.yu5@is.naist.jp
    }
    \and
    \IEEEauthorblockN{Ismail Arai, Masatoshi Kakiuchi, Arata Endo, Kazutoshi Fujikawa}
    \IEEEauthorblockA{
        \textit{Information iniTiative Center} \\ 
        \textit{Nara Institute of Science and Technology} \\ 
        Nara, Japan \\
        \{ismail, masato, endo.arata, fujikawa\}@itc.naist.jp
    }
}

% Figure省略する（Fig.）
  % cleveref 

\maketitle
\input{01-Abstract.tex}

\begin{IEEEkeywords}
Deep Learning, LSTM, Intelligent Transport Systems, Bus Ridership Prediction
\end{IEEEkeywords}

\input{02-Introduction.tex}
% 数便先の予測
% Therefore, a highly accurate method of predicting the number of passengers several bus services ahead is required to improve convenience for bus users.

% 数単位先の予測
% several units ahead

% 乗客数
% the number of passengers

% 特徴量
% データを変換することで生成される。
%  1つだけ
%   the feature
%  複数
%   the features

% 曜日
% a day of the week

% 時間帯
% time of day

% 章
% section

% 便数
% service number
%   bus service number

% Aという単位で
% in units of A

% それぞれのデータに data と明示的につけるべき?
% 

% 「バスで」はつけた方が良い?
% on a bus

% 「過去に」はつけた方が良い気がする。
% in the past

% It does not consider the relationship between the number of passengers at bus stops.

% サービス
% bus service

% 数便先
% several bus services

% 一般名詞の複数形
% 無冠詞で良い。

\input{03-RelatedWorks.tex}

\input{04-Data.tex}

\input{05-ProposedMethod.tex}

\input{06-Evaluation.tex}

\input{07-ResultAndDiscussion.tex}

\input{08-Conclusion.tex}

\input{09-Acknowledgment.tex}

% 適切な単語かどうかを確認
% perceive
% research or study → 一旦study
% 

% 冠詞
% day of the week は冠詞が必要かどうか

% 問題点
% It does not consider the relationship between the number of passengers at bus stops.
% In separate related studies, the number of passengers on buses in the past, time of day, day of the week, weather, and precipitation are each shown to be useful features.
% However, a prediction has yet to be made considering all these features.

% 用語の統一
% design 
% we designed a single LSTM-based model for all bus stops on one route.
% They designed a single LSTM-based model for all bus stops on multiple routes.

\bibliographystyle{IEEEtran}
\bibliography{paper.bib}

\end{document}

%% file: 01-Abstract.tex
\begin{abstract}
% 近年、人々にとって公共交通機関は必須のものである。
% 人々がバスに乗車するかを選択する際には、バス車内の混雑度が影響することが明らかである。
% そのため、バス利用者の利便性を向上するためには、バスの乗車人数を利用者に提供する必要がある。
% しかし、不正確な情報の提供はネガティブな体験を提供することになる懸念もある。
% このような背景から精度の高い予測をバス利用者に提供する必要がある。
% 多くの研究者が、これに関する研究をしている。
% しかし、関連研究をまとめると2つの課題がある。
% 1つ目が、連続するバス停間における乗車人数の関係性を考慮した予測をする必要があるが、それが行われていない点である。
% 2つ目が、関連研究でそれぞれ有効と示されている特徴量を全て使用した予測をしていない点である。
% 本研究では、この二つの問題点を解決する予測手法を提案する。
% 1つ目に対しては、バス停毎にLSTMを用い、モデルとしてはバス停全体で1つとするアーキテクチャを作成することで解決する。
% 2つ目に対しては、有効な全ての特徴量を入力することで解決する。
% 提案手法は既存手法と比較して最大でRMSEが27%向上した。
% そのため、提案手法が有効である。
Public transportation has been essential in people's lives in recent years.
Bus ridership is a factor in people's choice to board the bus.
Therefore, from the perspective of improving service quality, it is important to inform passengers who have not boarded the bus yet about future bus ridership.
However, there is a concern that providing inaccurate information may cause a negative experience.
Against this backdrop, there is a need to provide bus passengers who have not boarded yet with highly accurate predictions.
Many researchers are working on studies on this.
However, two issues summarize related studies.
The first is that the correlation of bus ridership between consecutive bus stops should be considered for the prediction.
The second is that the prediction has yet to be made using all of the features shown to be useful in each related study.
This study proposes a prediction method that addresses both of these issues.
We solve the first issue by designing an LSTM-based architecture for each bus stop and a single model for the entire bus stop.
We solve the second issue by inputting all useful data, the past bus ridership, day of the week, time section, weather, and precipitation, as features.
Bus ridership at each bus stop collected from buses operated by Minato Kanko Bus Inc, in Kobe city, Hyogo, Japan, from October 1, 2021, to September 30, 2022, were used to compare accuracy.
The proposed method improved RMSE by 23\% on average and up to 27\% compared to existing methods.
\end{abstract}

%% file: 02-Introduction.tex
\section{Introduction}
\label{Introduction}
Public transportation is closely tied to people's lives.
A local bus is one of the typical public transportation, but people perceive it as unreliable and congested\cite{bordagaray_modelling_2014}.
Therefore, from the perspective of improving service quality, it is important to inform passengers who have not boarded the bus yet about future bus ridership.
In fact, it is known that the number of vacant seats and the congestion of a bus affect the human's decision on whether to use the bus\cite{echaniz_spatial_2022,stradling_passenger_2007,atombo_indicators_2021,kim_passenger_2009}.
However, there is a concern that the inaccurate prediction of future bus ridership degrades the service quality of buses\cite{drabicki_modelling_2021}.
For this reason, a high-accuracy prediction method for bus ridership is required.

% Many researchers have studied the prediction for bus ridership\cite{yamamura,ouyang_lstm-based_2020,xu_lstm_2021,halyal_forecasting_2022,han_short-term_2019,karlaftis_statistical_2011,arabghalizi_data-driven_2020}.
% Without machine learning, bus ridership can be predicted based on the past bus ridership on each day of the week and time section\cite{ma_predicting_2014}.
% Machine learning can make bus ridership prediction more accurate than the statistical method by using various features other than the day of the week and time section\cite{halyal_forecasting_2022}.
% However, these predictions do not achieve 100\% of prediction accuracy.
% This is because the features are not well designed.

Many researchers have studied the prediction of bus ridership\cite{yamamura,ouyang_lstm-based_2020,xu_lstm_2021,halyal_forecasting_2022,han_short-term_2019,karlaftis_statistical_2011,arabghalizi_data-driven_2020}.
Statistical methods using historical bus ridership data, such as averages for each day of the week and time section\cite{ma_predicting_2014}, and machine learning methods using various data other than day of the week and time section have been used to predict bus ridership.
Among these methods, the latter can predict bus ridership with higher accuracy.
Furthermore, among the machine learning methods, the LSTM-based method, which allows input as time series data, is the most advanced in bus ridership prediction\cite{halyal_forecasting_2022}.
However, their accuracies are not 100\%, and there were issues for improvement.

To summarize the related work, there were two issues with feature design.
First, the correlation of bus ridership between consecutive bus stops has not been considered.
Second, there is no prediction method using all the useful features, the past bus ridership, day of the week, time section, weather, and precipitation.
This paper proposes a bus ridership prediction method for one bus service ahead with features, day of the week, time section, weather, precipitation, and bus ridership trend fed into multiple LSTM.
The evaluation results show that the proposed method improves RMSE compared to the existing method by 23\% on average and up to 27\%.

%% file: 03-RelatedWorks.tex
\section{Related work and issues}
\label{RelatedStudies}
This section reviews related work on a machine learning-based prediction for bus ridership.
After that, we describe their issues.

In our previous study, we tackled bus ridership prediction in service number units\cite{yamamura}.
We designed a model to input the ridership at all bus stops on one route in one LSTM layer.

Ouyang et al. aggregated ridership at each bus stop by 30-minute and predicted it several units ahead\cite{ouyang_lstm-based_2020}.
They designed a model to input the features for all bus stops on multiple routes in one LSTM layer.
The features are the past bus ridership, day of the week, time section, bus stop number, latitude, longitude, and labels for the 17 classifications of facilities.

Xu et al. predicted ridership for each bus route\cite{xu_lstm_2021}.
They aggregated ridership per bus route into hourly units and predicted it several units ahead.
In this method, a model exists for each bus route.
The features used by this model are the past bus ridership, day of the week, and weather.

\begin{figure}[t]
  \centering
  \includegraphics[width=0.9\linewidth]{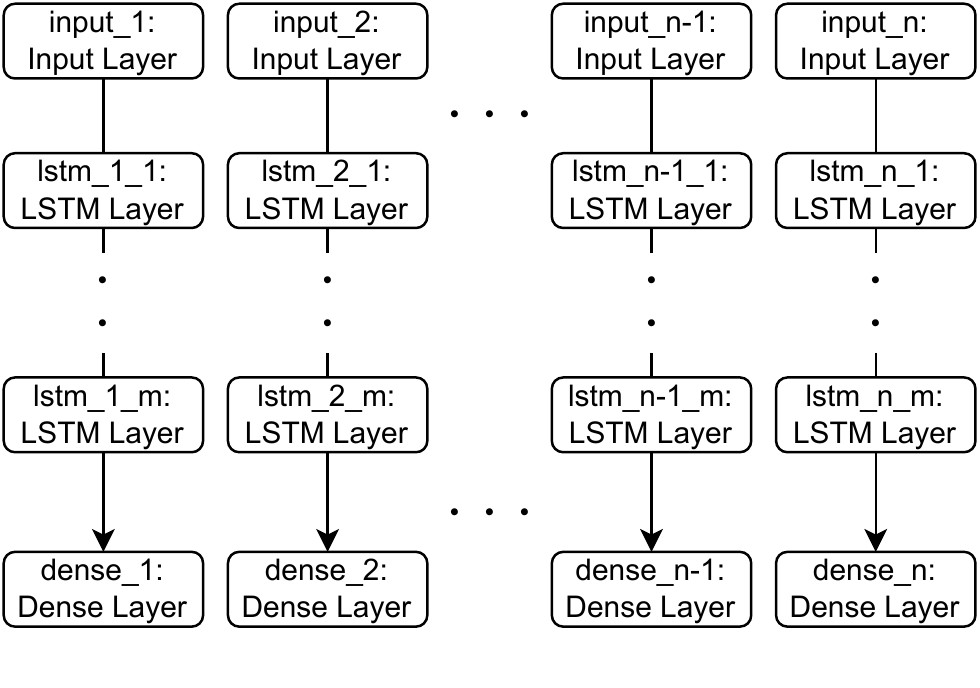}
  \caption{Architecture of Halyal's bus ridership prediction method\cite{halyal_forecasting_2022}.}
  \label{fig:existing_architecture}
\end{figure}

Halyal et al. aggregated ridership at each bus stop into four units (15, 30, 45, and 60 minutes) and predicted it several units ahead\cite{halyal_forecasting_2022}.
This method requires a model for each bus stop, respectively, as shown in Fig. \ref{fig:existing_architecture}.
For example, when there are five bus stops on a route, this method requires five models (n=5).
The feature is the past bus ridership.

Han et al. predicted ridership in units of service number\cite{han_short-term_2019}.
In this study, their proposed method adopts an architecture where a model exists for each bus stop.
Their proposed method uses a past bus ridership as a feature.
Furthermore, they showed that their proposed method improves prediction accuracy by combining two optimization algorithms, Nadam\cite{dozat_incorporating_2016} and SGD\cite{noauthor_stochastic_1951}, for training.

All of these prediction models are based on LSTM.
However, they can be categorized from spatial and temporal perspectives, respectively.\\

\begin{description}
  \item[Spatial perspective]\mbox{}\\These models are categorized into a per-bus-stop model (We\cite{yamamura}, Ouyang et al.\cite{ouyang_lstm-based_2020}, Halyal et al.\cite{halyal_forecasting_2022}, and Han et al.\cite{han_short-term_2019}) and a per-bus-route model (Xu et al.\cite{xu_lstm_2021}). 
  The former model predicts ridership at the time of bus stop departure.
  The latter model predicts ridership on a bus route.
  The former model is more convenient for passengers than the latter model.\\
  \item[Temporal perspective]\mbox{}\\These models are categorized into a per-service model (We\cite{yamamura}, Han et al.\cite{han_short-term_2019}) and a per-aggregated-unit model (Ouyang et al.\cite{ouyang_lstm-based_2020}, Xu et al.\cite{xu_lstm_2021}, and Halyal et al.\cite{halyal_forecasting_2022}).
  The former model predicts ridership by service number.
  The latter model predicts ridership by units aggregated at specific times.
  The former model is more convenient for passengers than the latter model.
  \\
\end{description}

These studies have the following two issues.
First, the correlation of bus ridership between consecutive bus stops has not been considered.
If there is a correlation in ridership at the bus stops, the input design should consider it to improve prediction accuracy.
The study by Xu et al. predicted it by bus route unit.
However, the correlation of bus ridership at the bus stops cannot be considered when predicting ridership by bus route unit.
Learning could be more complicated if ridership for all bus stops were input as one-dimensional data in a time series, as in our previous study.
Because one LSTM layer needs to find the correlation between ridership in service number and at bus stops.
Although this method inputs ridership at all bus stops, it does not consider the correlation between bus stops so that the input design is not appropriate.

Second, bus ridership prediction using all useful features has yet to be addressed.
The useful features in bus ridership prediction from related studies are the past bus ridership, day of the week, time section, weather, precipitation, bus stop numbers, latitude, longitude, and 17 different labels for facilities.
Among these, bus stop numbers, latitude, longitude, and 17 different labels for facilities are not available in our approach that adopts a per-bus-stop and per-service prediction model.
Therefore, the useful features for our approach are the past bus ridership, day of the week, time section, weather, and precipitation.
Since it is verified that each feature contributes to the accuracy improvement of bus ridership predictions, there is a possibility of improving the prediction accuracy by using all the useful features.

Therefore, in this paper, we tackle the following two issues:\\
\begin{quote}
  \begin{itemize}
  \item[1.] The correlation of bus ridership between consecutive bus stops has not been considered.\\
  \item[2.] There is no prediction method using all the useful features, the past bus ridership, day of the week, time section, weather, and precipitation.
  \\
\end{itemize}
\end{quote}

%% file: 04-Data.tex
\section{Dataset}
\label{Data}
This section describes how to collect data for this study: bus ridership, day of the week, time section, weather, and precipitation data.

\subsection{Bus ridership data}
\begin{figure}[t]
  \centering
  \includegraphics[width=0.95\linewidth]{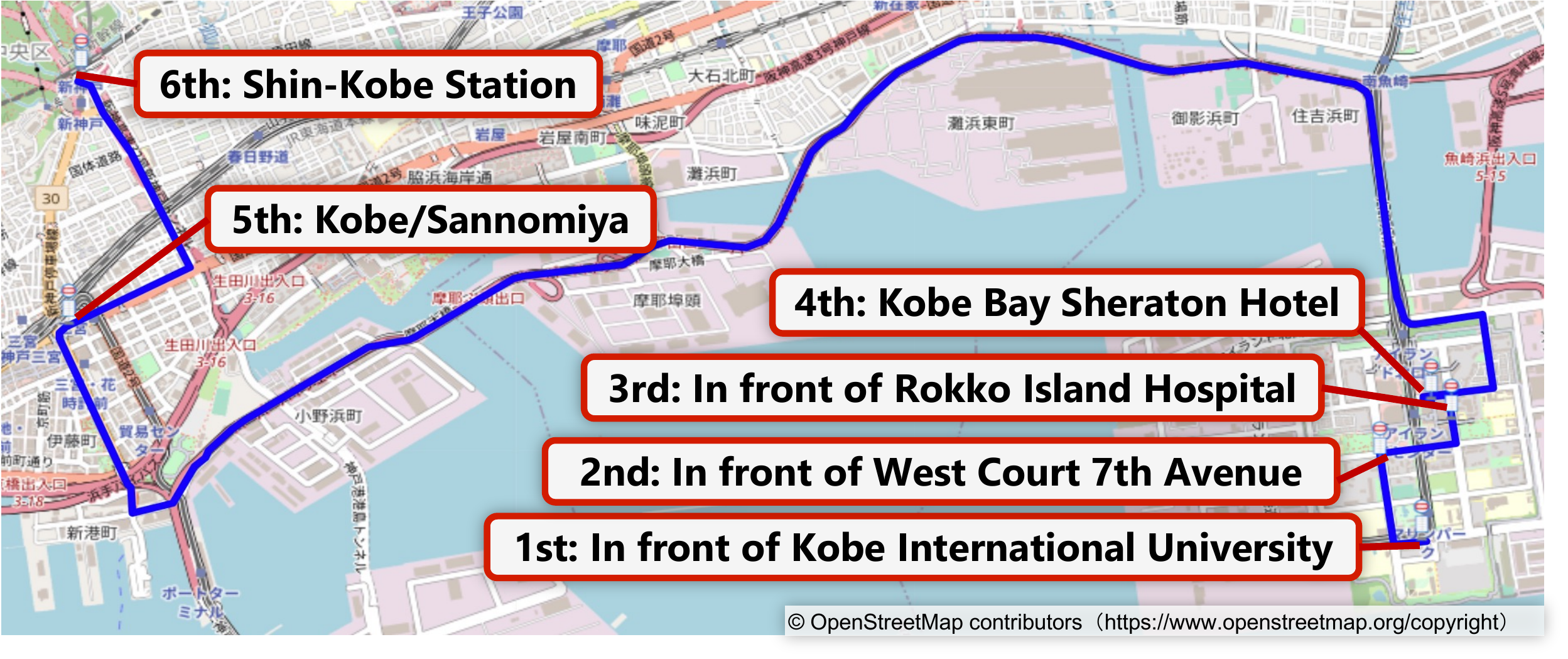}
  \caption{Map of a route handled in this study. The bus stop names are written in the ``order'' in which the buses arrive.}
  \label{fig:Map}
\end{figure}

% 1. バス路線情報
We collected bus ridership data needed for this study from buses operated by Minato Kanko Bus Inc.
There are two types of bus service, looped operations, and inbound/outbound operations.
Since the latter type of operation is common in Japan, experiments were conducted using the latter type.
In this paper, the bus stop names are written in the ``order'' in which the buses arrive.
We also used inbound operation data from the 1st bus stop to the last bus stop out of inbound/outbound, as shown in Fig. \ref{fig:Map}.
This route is located in Kobe city, Hyogo, Japan, with six bus stops.
On this route, buses leave the 1st bus stop every 30 minutes in the morning and evening when ridership is high and every 60 minutes at other times when ridership is low.

\begin{figure}[t]
  \centering
  \includegraphics[width=1.0\linewidth]{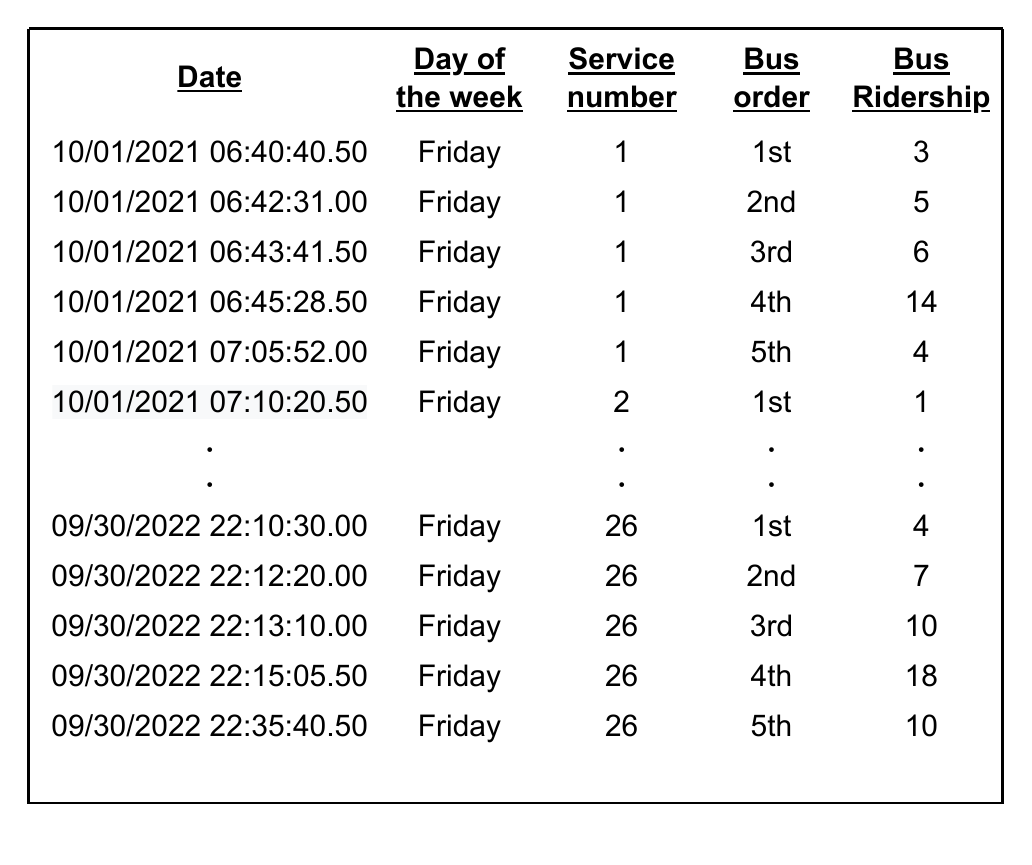}
  \caption{The data structures that are processed from the collected data and handled in this study.}
  \label{fig:Data_structure}
\end{figure}

\begin{figure}[t]
  \centering
  \includegraphics[width=1.0\linewidth]{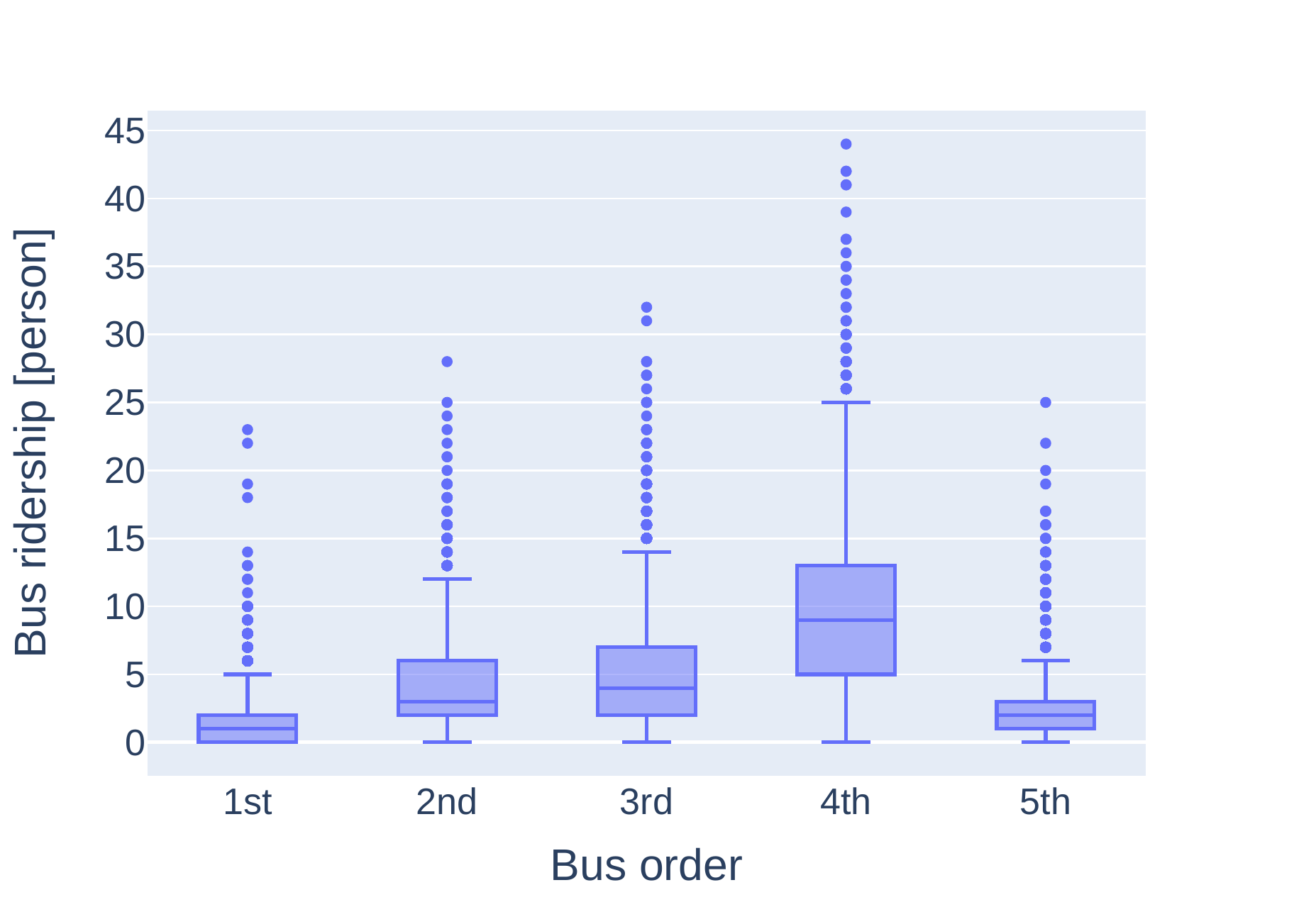}
  \caption{Box plot of ridership at each bus stop.}
  \label{fig:boxplot}
\end{figure}

% 2. 生データ収集
% 3. 加工したデータ
The data periods were from October 1, 2021, to September 30, 2022.
In this study, we use a per-bus-stop model to predict.
Therefore the prediction target is ridership at the time of departure from the bus stop, data from five bus stops, excluding the terminal, are used.
The data structure collected and processed from the buses handled in this study is shown in Fig. \ref{fig:Data_structure}.
We release the data, which is available on a GitHub repository\footnote{\url{https://github.com/inet-lab-naist/Minato_bus_ridership_data}}.
Fig. \ref{fig:boxplot} shows a box plot of bus ridership at each bus stop.
The 4th bus stop has the highest ridership, and the 1st bus stop has the lowest. 
Also, the 4th bus stop has a high variance.

% \begin{table}[t]
%   \centering
%   \caption{The bus stop names are written in the ``order'' in which the buses arrive.}
%   \label{fig:bus_stops}
%   \scalebox{1.2}{
%     \begin{tabular}{rl}
%       \hline \hline
%         Order & Bus stop name \\
%         \hline
%         1st & In front of Kobe International University \\
%         2nd & In front of West Court 7th Avenue \\
%         3rd & In front of Rokko Island Hospital \\
%         4th & Kobe Bay Sheraton Hotel \\
%         5th & Kobe/Sannomiya \\
%         \hline
%       \end{tabular}
%     }
% \end{table}

\subsection{Weather and precipitation data}
Weather and precipitation data are published by the Japan Meteorological Agency.
Hourly weather and precipitation data from 6:00 to 23:00 from October 1, 2021, to September 30, 2022, based on the location of the route to be evaluated.
Since the first bus leaves at 6:40 and the last arrives at 22:45, we obtained data from 6:00 to 23:00.
Table \ref{fig:weather} is a breakdown of the weather data obtained.
In the proposed method described in section \ref{ProposedMethod}, the acquired data are formatted into two types since they are classified as whether it is raining or not.
Sunny and Cloudy are defined as weather without rain.
Rain showers, Rain, Freezing rain and Other which is precipitation but not classifiable are defined as weather with rain.
The number of observations is summarized as 5580 for no rain and 618 for rainy weather.
% TODO：以下の定義
\begin{table}[t]
    \centering
    \caption{Hourly weather data from 6:00 to 23:00 from October 1, 2021, to September 30, 2022, in Kobe city, Hyogo, Japan.}
    \label{fig:weather}
    \scalebox{1}{
      \begin{tabular}{cr}
        \hline \hline
          Weather & Number of observations \\
          \hline
          Sunny & 3308 \\
          Cloudy & 2272 \\
          \hline
          Rain showers & 539 \\
          Rain & 12 \\
          Freezing rain & 54 \\
          Other & 13 \\
          \hline
        \end{tabular}
    }
  \end{table}

%% file: 05-ProposedMethod.tex
% \section{A bus ridership prediction method considering the time section, weather, and the ridership correlation between bus stops}
\section{Bus ridership prediction with time section, weather, and ridership trend aware \\ multiple LSTM}
\label{ProposedMethod}
% 今回の提案では既存研究と同じLSTMを採用し、その上で特徴量の設計を工夫した。
This section describes the proposed method to solve the two issues addressed in section \ref{RelatedStudies}.
We solve the first issue by providing an LSTM layer at each bus stop and building an architecture that provides one model for all bus stops (defined as Multiple-LSTM model).
Also, we solve the second issue by inputting the past bus ridership, day of the week, time section, weather, and precipitation into an LSTM layer (defined as K-feature-integrated model).
Finally, we propose a method that combines these two models.

\subsection{Multiple-LSTM model}
First, the correlation coefficient of ridership between bus stops was determined to see if there was a correlation between ridership at the bus stops.
Table \ref{fig:Correlation} shows that the minimum correlation coefficient between bus stops was 0.3104, and the maximum was 0.9208.
Therefore, we considered that a correlation existed between each other at the five bus stops.

\begin{figure}[t]
  \centering
  \includegraphics[width=1\linewidth]{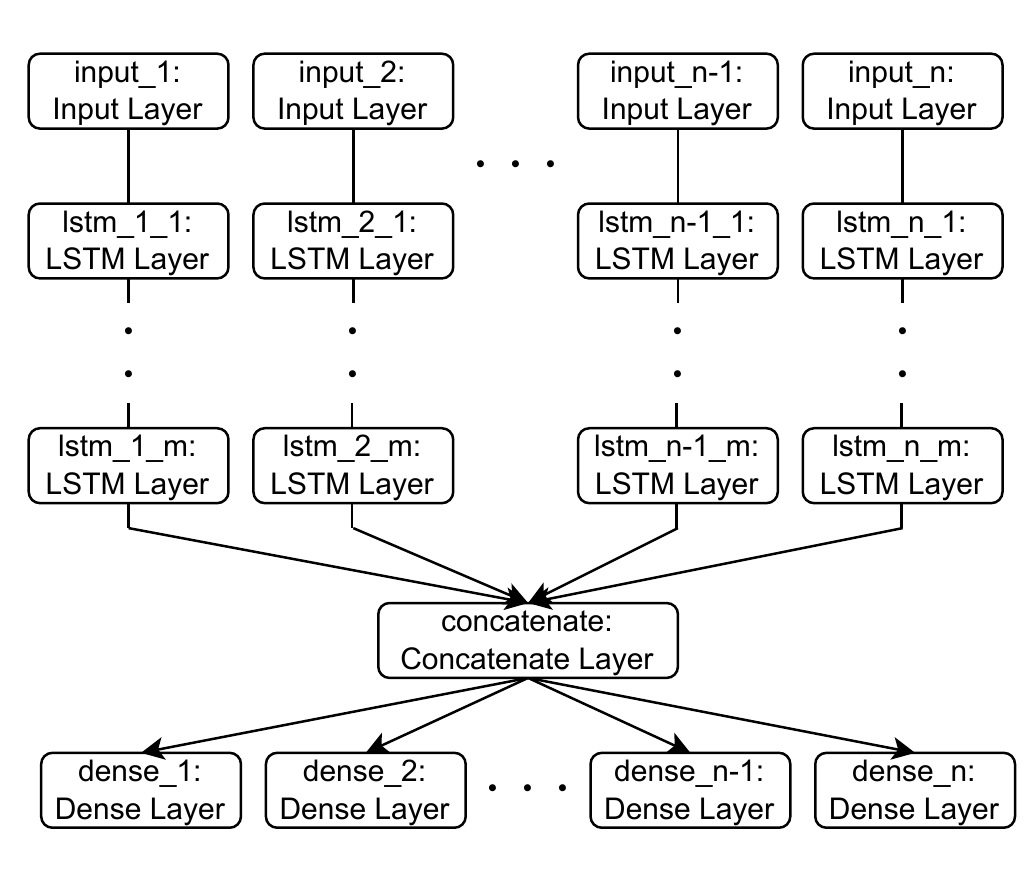}
  \caption{Architecture of the Multiple-LSTM model. An LSTM is designed for each bus stop and only one model is designed for all bus stops. If there are five target bus stops, n is 5.}
  \label{fig:proposed_architecture}
\end{figure}

Designing a separate model for each bus stop (Fig. \ref{fig:existing_architecture}) would not allow predictions considering the correlation among bus stops.
We also considered that if one model that includes one LSTM predicts ridership at all bus stops, it would be difficult to find correlation among bus stops during training.
Therefore, an LSTM layer should be provided for each bus stop, and there should be only one model for all bus stops.
The model architecture to meet these requirements is shown in Fig. \ref{fig:proposed_architecture}.
The past ridership at each bus stop is fed into the Input Layer, and ridership one service ahead of each bus stop is predicted in the Dense Layer.

\begin{table}[t]
  \centering
  \caption{The correlation coefficient of ridership between bus stops.}
  \label{fig:Correlation}
  \scalebox{1}{
    \begin{tabular}{rccccc}
      \hline \hline
        \multicolumn{1}{l|}{} & 1st &  2nd & 3rd & 4th & 5th                 \\
        \hline
        \multicolumn{1}{l|}{1st}        & -  & -  & - & - & -       \\
        \multicolumn{1}{l|}{2nd}        & 0.4678  & -  & - & - & - \\
        \multicolumn{1}{l|}{3rd}        & 0.3644  & 0.9208  & - & - & - \\
        \multicolumn{1}{l|}{4th}        & 0.3104  & 0.7646  & 0.8339 & - & - \\
        \multicolumn{1}{l|}{5th}        & 0.4011  & 0.5269  & 0.5345 & 0.5942 &  -         \\
        \hline
      \end{tabular}
  }
\end{table}

\subsection{K-feature-integrated model}
We input all data as features that are useful in related work into LSTM.
As mentioned in section \ref{Data}, the number of buses in this study varies by time of day.
Therefore, instead of using the time section as a feature, a value indicating the service number was used as the feature.
In addition, we used a flag of precipitation as a feature instead of weather according to Xu et al.\cite{xu_lstm_2021}.
As a result, we used the past bus ridership, day of the week, service number, rainfall flag, and precipitation as features.
The past bus ridership and precipitation are Min-Max normalized.
Service number, day of the week, and rainfall flag are one-hot encoded.
In other words, we convert the features to 37 dimensions.
When learning with deep learning, it is common to normalize quantitative variables and one-hot-encode qualitative variables.

\begin{figure}[t]
    \centering
    \includegraphics[width=0.95\linewidth]{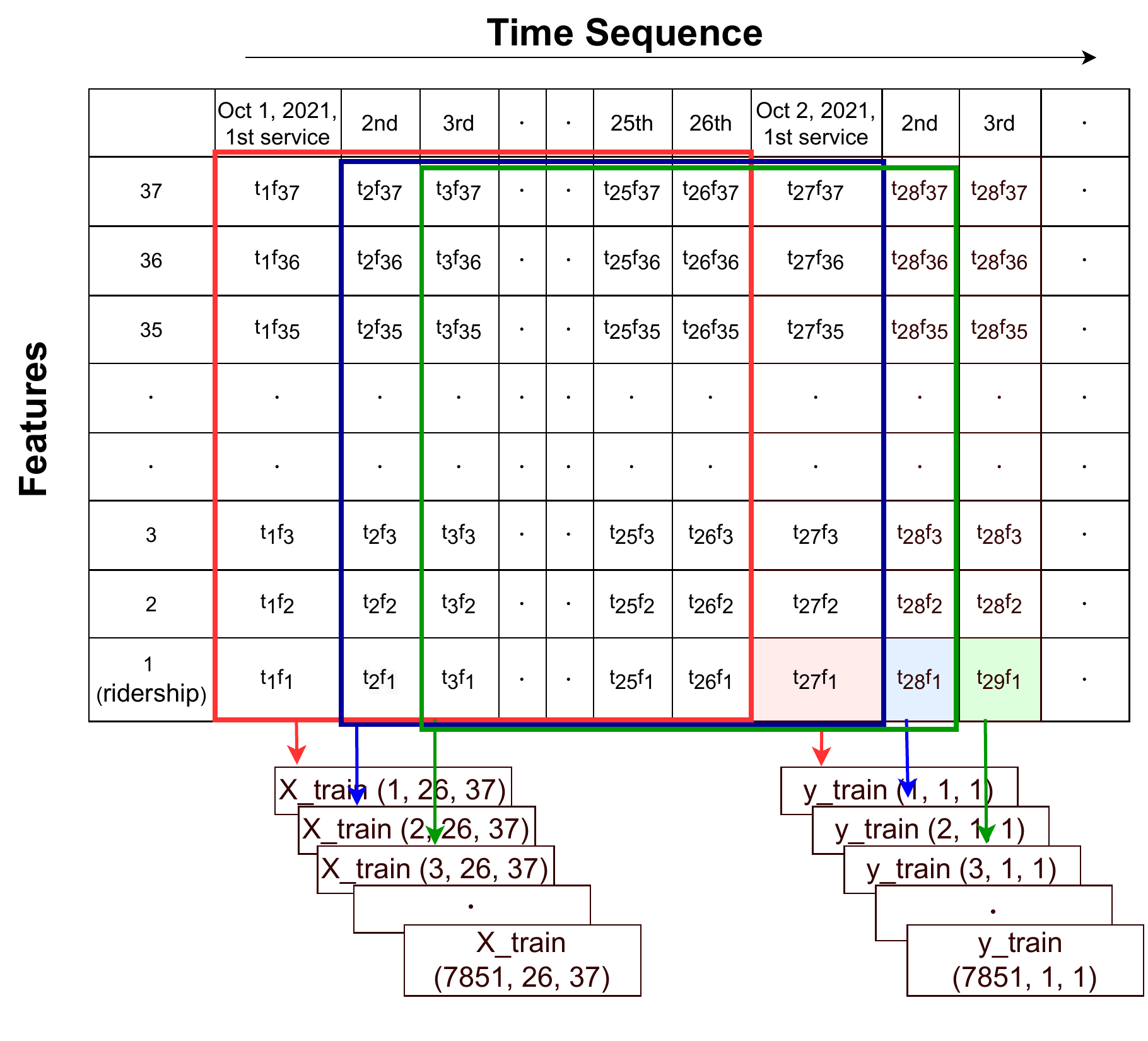}
    \caption{Dataset splitting process of the K-feature-integrated model. In this case, the number of look-back is 26.}
    \label{fig:DataSplitDetail}
  \end{figure}

Fig. \ref{fig:DataSplitDetail} shows the process of splitting the dataset to learn.
It is a case of predicting ridership one service ahead based on data on ridership, service number, day of the week, rainfall flag, and precipitation for the previous 26 services.
The 26 is the number of look-back services and tuned value.
This tuning is discussed in section \ref{Evaluation}.
X\_train and y\_train represent the input and output during training, respectively.
By sliding one service at a time, we obtain the data structure of the input (index, the number of look-back services, the number of features) and the corresponding output (index, 1, bus ridership).
We call this the 5-feature-integrated model among the K-feature-integrated model.
It should be noted that Fig. \ref{fig:DataSplitDetail} shows the process of splitting a dataset at one bus stop.
Since five bus stops are targeted in this case, this splitting process is conducted at each of the five bus stops, and each split item is input to the Input Layer shown in Fig. \ref{fig:proposed_architecture}, respectively.

%% file: 06-Evaluation.tex
\section{Evaluation}
\label{Evaluation}
This section describes an experiment to evaluate whether a method that addresses the problems proposed in section \ref{ProposedMethod} improves prediction accuracy.

\begin{table*}[t]
  \centering
  \caption{Details of each method for use in the evaluation.}
  \label{fig:Details_of_each_method}
  \scalebox{1}{
    \begin{tabular}{rl}
      \hline \hline
      Method & Detail \\ \hline
      Method A & Multiple-LSTM model with the past bus ridership as a feature.\\
      Method B & \begin{tabular}[l]{@{}l@{}}Multiple-LSTM model \& 3-feature-integrated model \\with the past bus ridership, day of the week, and service number as features.\end{tabular}\\
      Method C & \begin{tabular}[l]{@{}l@{}}Multiple-LSTM model \& 3-feature-integrated model \\with the past bus ridership, rainfall flag, and precipitation as features.\end{tabular}\\
      Method D & \begin{tabular}[l]{@{}l@{}}Multiple-LSTM model \& 5-feature-integrated model \\with the past bus ridership, day of the week, service number, \\rainfall flag, and precipitation as features.\end{tabular}\\
      Halyal's method & Design a model for each bus stop with the past bus ridership as a feature. \\ 
      Statistical method & The average values classified by the bus stop and service number are the predicted values.\\ \hline
    \end{tabular}
  }
  \end{table*}

\subsection{Evaluation method}
\label{Evaluation_method}
% 書く順番を考える。
% まずは評価するものを書く必要がある。
We evaluate the six methods shown in Table \ref{fig:Details_of_each_method} to confirm if the Multiple-LSTM model, K-feature-integrated model, and their combinations improve prediction accuracy.
Method A is the Multiple-LSTM model.
Method D is a combination of the Multiple-LSTM model and the 5-feature-integrated model.
Halyal's method requires a model for each bus stop, respectively, as shown in Fig. \ref{fig:existing_architecture}.
Therefore, it was employed as a comparison to evaluate the effect of the Multiple-LSTM model.
Methods B and C use some of the features.
Therefore, it was employed as a comparison to evaluate the effect of the 5-feature-integrated model.
To compare the accuracy of the statistical prediction, we employed Statistical method with average ridership classified by the bus stop and service number for October 1, 2021, to August 31, 2022, which was used as the prediction value.

We split our data into training data from October 1, 2021, to July 31, 2022, validation data from August 1, 2022, to August 31, 2022, and test data from September 1, 2022, to September 30, 2022.
Since the data used are time series data, to prevent leakage, the validation data should be from a later period than the training data period, and the test data should be from an even later period.
In general, the division of training, validation, and test data should be about 80\%, 10\%, and 10\%, respectively.
In this experiment, prediction accuracy is compared using RMSE.
In this evaluation method, a smaller number indicates higher accuracy.

\subsection{Parameter tuning}
\label{Parameter_tuning}
Both Method D which is the proposed method, and Halyal's method used Hyperband\cite{li_hyperband_nodate} for parameter tuning.
Hyperband is a Random Search\cite{bergstra_random_nodate} that incorporates early termination and adaptive allocation of computational resources.

Table \ref{fig:hyperband_result} shows the candidate values and results of parameter tuning for Method D and Halyal's method.
The candidate values for each parameter are the values used in tuning the parameters of the neural network.
The proposed method designs one model with five bus stops, which results in one parameter tuning at all bus stops.
On the other hand, since Halyal's method designs a model for each bus stop, parameter tuning was also conducted at each bus stop.
The number of LSTM layers represents the number of m in Fig. \ref{fig:existing_architecture} and Fig. \ref{fig:proposed_architecture}, which is from 1 to 3.
The sequence length was set to 26 or 182, considering the periodicity of a day or a week, since the routes in the data used in this study have 26 bus services per day.

\begin{table*}[t]
    \centering
    \caption{Candidate values and best results of parameter tuning using Hyperband for Method D and Halyal's method.}
    \label{fig:hyperband_result}
    \scalebox{1}{
      \begin{tabular}{cccccccc}
        \hline \hline
        \multirow{2}{*}{Hyperparameter type} & \multirow{2}{*}{List of values} & \multirow{2}{*}{\begin{tabular}[c]{@{}c@{}}Best result for \\method D\end{tabular}} & \multicolumn{5}{l}{Best result per bus stop for Halyal's method} \\ \cline{4-8}
        & & & \multicolumn{1}{c}{\begin{tabular}[c]{@{}l@{}}1st\end{tabular}} & \multicolumn{1}{c}{\begin{tabular}[c]{@{}l@{}}2nd\end{tabular}} & \multicolumn{1}{c}{\begin{tabular}[c]{@{}l@{}}3rd\end{tabular}} & \multicolumn{1}{c}{\begin{tabular}[c]{@{}l@{}}4th\end{tabular}} & \multicolumn{1}{c}{\begin{tabular}[c]{@{}l@{}}5th\end{tabular}} \\ \hline
        Batch size & 16, 32, 64, 128, 256 & \multicolumn{1}{S}{16} & \multicolumn{1}{S}{256} & \multicolumn{1}{S}{32} & \multicolumn{1}{S}{128} & \multicolumn{1}{S}{16} & \multicolumn{1}{S}{128} \\
        Sequence length & 26, 182 & \multicolumn{1}{S}{26} & \multicolumn{1}{S}{26} & \multicolumn{1}{S}{26} & \multicolumn{1}{S}{26} & \multicolumn{1}{S}{26} & \multicolumn{1}{S}{26} \\
        LSTM nodes & 16, 32, 64, 128, 256 & \multicolumn{1}{S}{64} & \multicolumn{1}{S}{16} & \multicolumn{1}{S}{128} & \multicolumn{1}{S}{32} & \multicolumn{1}{S}{32} & \multicolumn{1}{S}{16} \\
        Number of LSTM layers & 1, 2, 3 & \multicolumn{1}{S}{1} & \multicolumn{1}{S}{1} & \multicolumn{1}{S}{3} & \multicolumn{1}{S}{3} & \multicolumn{1}{S}{3} & \multicolumn{1}{S}{1} \\
        Learning rate & 0.01, 0.001, 0.0001 & \multicolumn{1}{S}{0.001} & \multicolumn{1}{S}{0.01} & \multicolumn{1}{S}{0.01} & \multicolumn{1}{S}{0.01} & \multicolumn{1}{S}{0.01} & \multicolumn{1}{S}{0.01} \\ 
        Optimizers & \begin{tabular}[c]{@{}c@{}}Adadelta, Adagrad, Adam, \\Adamax, Ftrl, Nadam, \\RMSprop, SGD\end{tabular} & \multicolumn{1}{c}{Adam} & \multicolumn{1}{c}{RMSprop} & \multicolumn{1}{c}{Nadam} & \multicolumn{1}{c}{Nadam} & \multicolumn{1}{c}{RMSprop} & \multicolumn{1}{c}{Nadam} \\ \hline
      \end{tabular}
    }
    \end{table*}

%% file: 07-ResultAndDiscussion.tex
\section{Result and discussion}
\label{ResultAndDiscussion}
This section describes the results of the evaluation experiments in section \ref{Evaluation} and discusses them.

Table \ref{fig:result} shows the evaluation results.
The best RMSE on each bus stop is bold.
The results of Method A and Halyal's method show that designing one model for all bus stops, with an LSTM layer at each bus stop, is more accurate than designing a model for each bus stop.
This method is effective when the bus route has a correlation coefficient in the value of ridership between bus stops, as shown in Table \ref{fig:Correlation}.

From the results of Methods B, C, and D, the method that considers bus ridership, the day of the week, service number, rainfall flag, and precipitation, all input as features, has better accuracy than the method that does not consider them.
The results of Methods A and C show that the method that considers rainfall flag and precipitation are more accurate at the 1st and the 2nd bus stops.
However, Method A is more accurate at the 3rd, 4th, and 5th bus stops.
Therefore, it is clear that adding the rainfall flag and precipitation as features does not improve accuracy at some bus stops.
Comparing the results of Methods B and D, Method D is more accurate at all bus stops.
The accuracy was improved by the interaction of not only the rainfall flag and precipitation but also other data, such as the day of the week and service number, which were input as features simultaneously.

Comparing the RMSE values between bus stops shows that the 4th bus stop has the worst accuracy.
This is because, as shown in Fig. \ref{fig:boxplot}, the variance of ridership at the 4th bus stop is high and difficult to predict.

% Comparing the results of Method D and Halyal's method, the combination of the Multiple-LSTM model and the K-feature integration model was more effective than Halyal's methods, with a 27\% improvement in accuracy at Bus Stop 4.
Comparing the results of Method D and Halyal's method, the combination of the Multiple-LSTM model and the K-feature integration model was more effective than Halyal's method, improving accuracy up to 27\%, 21\%, 16\%, 27\%, and 25\% at Bus Stop 1, 2, 3, 4, and 5, respectively.

\begin{table}[t]
    \centering
    \caption{Evaluation results at five bus stops using RMSE for each method.}
    \label{fig:result}
    \scalebox{1}{
      \begin{tabular}{rlllll}
        \hline \hline
        \multirow{2}{*}{} & \multicolumn{5}{l}{RMSE {[}person{]}} \\ \cline{2-6}
        & \multicolumn{1}{l}{\begin{tabular}[c]{@{}l@{}}1st\end{tabular}} & \multicolumn{1}{l}{\begin{tabular}[c]{@{}l@{}}2nd\end{tabular}} & \multicolumn{1}{l}{\begin{tabular}[c]{@{}l@{}}3rd\end{tabular}} & \multicolumn{1}{l}{\begin{tabular}[c]{@{}l@{}}4th\end{tabular}} & \begin{tabular}[c]{@{}l@{}}5th\end{tabular} \\ \hline
        Method A & \multicolumn{1}{l}{1.419} & \multicolumn{1}{l}{2.399} & \multicolumn{1}{l}{2.716} & \multicolumn{1}{l}{4.056} & 1.963 \\
        Method B & \multicolumn{1}{l}{1.311} & \multicolumn{1}{l}{2.217} & \multicolumn{1}{l}{2.530} & \multicolumn{1}{l}{3.694} & 1.905 \\
        Method C & \multicolumn{1}{l}{1.357} & \multicolumn{1}{l}{2.377} & \multicolumn{1}{l}{2.781} & \multicolumn{1}{l}{4.164} & 2.020 \\
        Method D & \multicolumn{1}{l}{\textbf{1.128}} & \multicolumn{1}{l}{\textbf{2.122}} & \multicolumn{1}{l}{\textbf{2.459}} & \multicolumn{1}{l}{\textbf{3.636}} & \textbf{1.737} \\
        Halyal's method & \multicolumn{1}{l}{1.549} & \multicolumn{1}{l}{2.683} & \multicolumn{1}{l}{2.928} & \multicolumn{1}{l}{4.981} & 2.320 \\
        Statistical method & \multicolumn{1}{l}{1.389} & \multicolumn{1}{l}{2.474} & \multicolumn{1}{l}{3.047} & \multicolumn{1}{l}{4.517} & 2.032 \\ \hline
      \end{tabular}
    }
  \end{table}

% バス停1: (1.549-1.128)/1.549*100=27.178825048418336=27
% バス停2: (2.683-2.122)/2.683*100=20.909429742825196=21
% バス停3: (2.928-2.459)/2.928*100=16.017759562841526=16
% バス停4: (4.981-3.636)/4.981*100=27.00260991768721=27
% バス停5: (2.320-1.737)/2.320*100=25.129310344827577=25
% 5つのバス停の平均: 23.247586923319968

%% file: 08-Conclusion.tex
\section{Conclusion}
\label{Conclusion}
In order to improve the convenience of passengers who have not boarded the bus yet, a method that can predict ridership with several bus services ahead with high accuracy is needed.
Many researchers have studied the prediction of bus ridership.
However, there were two issues in related studies.
First, the correlation of bus ridership between consecutive bus stops has not been considered for the prediction yet.
Second, there is no prediction method using all the useful features, the past bus ridership, day of the week, time section, weather, and precipitation.
% この研究では上記二つの問題に取り組んだ話が必要
This study addressed the above two issues.
We solve the first issue by designing each LSTM at each bus stop and building an architecture that provides one model for all bus stops.
We solve the second issue by simultaneously inputting the past bus ridership, day of the week, service number, rainfall flag, and precipitation into LSTM as features.
This experiment evaluated the comparison on one service ahead prediction task.
We confirm that the combined method to solve the first and second issues is the most accurate of the six methods at all bus stops.
We improve RMSE by 23\% on average and up to 27\% compared to Halyal's methods.

There are three future works.
The first is to create a model that considers event data.
It is known that an event around bus stops, such as festivals, exhibitions, concerts, and workshops, tends to affect bus ridership.
However, related work has not proposed methods to account for it yet.
The second is to evaluate other bus routes.
This work is necessary to show that the proposed method is effective for other routes.
The third is to evaluate the ridership prediction for the next few days.
For the convenience of bus users, a prediction farther than one service ahead is necessary.

%% file: 09-Acknowledgment.tex
\section*{Acknowledgment}
This work was supported by JSPS KAKENHI Grant Numbers JP20K11789, JP20H04183.
The authors want to thank Minato Kanko Bus Inc., Japan, for providing the sensor data.